**Visualising the structure of architectural open spaces based on shape analysis[1]**

Sanjay Rana (corresponding author) and Mike Batty
Centre for Advanced Spatial Analysis
University College London
1-19 Torrington Place
London WC1E 6BT
UK
s.rana@ucl.ac.uk[2]
Phone 020-76791782
Fax 020-7813 2843

*Abstract*

This paper proposes the application of some well known two-dimensional geometrical shape descriptors for the visualisation of the structure of architectural open spaces. The paper demonstrates the use of visibility measures such as distance to obstacles and amount of visible space to calculate shape descriptors such as convexity and skeleton of the open space. The aim of the paper is to indicate a simple, objective and quantifiable approach to understand the structure of open spaces otherwise impossible due to the complex construction of built structures.

## 1. INTRODUCTION

Visualisation and interpretation of architectural spaces are intellectually challenging and multi-faceted exercises with fairly wide ranging applications and implications. In the context of urban planning, they are most commonly undertaken to evaluate the usage of the architectural space to ensure efficient navigation and accessibility [1]. These exercises clearly assume a certain influence of the built structures on the human cognition. However, what aspect of architectural space affects the human behaviour still remains an open debate. In this respect, it is closely similar to an exercise to identify the unknown visual variables in information visualization. Since a quantitative analysis of the architectural geometric structure on a large scale will be a daunting computational task, the open space bounded by the built structures is studied instead [2]. The study of architectural open spaces essentially involves the computation of the visibility polygon or isovist (space visible all around a viewpoint, see Figure 1a) from a viewpoint and calculating various shape measures of the visibility polygon. The isovist computation involves drawing rays from the viewpoint at very fine (<<0.1 degrees) equal angular intervals and collecting the distance to the obstacles. A summary of the entire open space can be obtained by performing the isovist computation at a number of densely located set of viewpoints (Figure 1b). For example, Figure 1c shows the variation in the amount of space visible from different parts the simple T-shape. A number of other shape measures such as convexity, clustering coefficient and radial measures are also computed to visualise the open space structure. See Batty and Rana [3] for more details on other isovist measures.

---

[1] Also published in International Journal of Architectural Computing, 2(1), 2004
[2] Now at Department of Geography, University of Leicester, Leicester LE1 7RH, UK, sr115@le.ac.uk

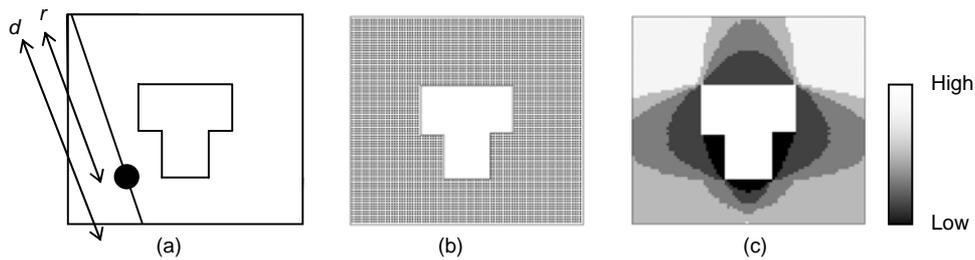

Figure 1: An open space with a T-shape object. (a) The isovist from a viewpoint and the radial (r) and diametric (d) lengths, (b) discretisation of the open space with a dense grid of viewpoints and, (c) area visible from viewpoints plotted as a field.

Another type of open space shape description is based upon the dominant lines of visibility, which are considered to act as subtle cues subconsciously followed by walkers. Hillier and Hanson [4] proposed the first such type of dominant lines of visibility called axial lines. The generation of axial lines involves the decomposition of the entire open space into non overlapping convex subspaces and drawing the minimum number of lines which could connect the subspaces (Figure 2)[3]. While, the calculation of axial lines for relatively simple architectural open space systems is easily done, it is non-trivial to compute an exact set of axial lines for larger spaces. The main shortcoming of this approach is that the space can be subdivided into more than one arrangement of convex subspaces as shown in Figure 2 in the simple case of the T-shape. Although we are able to produce a consistent set of three axial lines in this simple case, in practical examples, axial lines are very likely to vary with the designer.

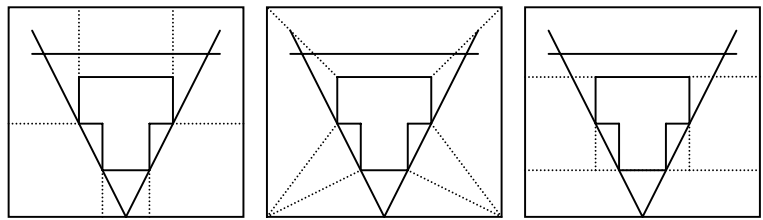

Figure 2: Partitioning of a single open space into many valid convex subspaces and the axial lines. Dotted lines indicate sub-space boundary and solid lines are axial lines.

---

[3] The discipline which deals with isovist analysis and axial lines in urban planning is called space syntax.

In this paper, we will demonstrate three methods for the derivation of the lines of dominant visibility, quite similar in properties to axial lines i.e. allow partition of open space into discrete connected sub-spaces, based entirely on shape measures. Therefore, our approach is novel and objective. The isovists and the other related measures were computed using *Isovist Analyst* developed by the first author, which is an extension to the popular Geographic Information System (GIS) software *ArcView*. *Isovist Analyst* is available free of charge for non-commercial and non-profit purposes by writing to the authors. The primary inputs of the analysis are 2D architectural plans e.g. of cities or building indoors either in GIS or CAD (e.g. DXF) formats and the layout of the viewpoints within the open space between the plans. *Isovist Analyst* can compute upto twenty measures related to the shape of the open space as seen from the viewpoints.

## 2. DOMINANT LINES OF VISIBILITY

Our approaches are based on two types of isovist measures namely the minimum radial length (MRL) and maximum diametric length (MDL) (see Figure 1a). In short, the minimum radial length of an isovist is the shortest distance to the boundary from a view point and the maximum diametric length is the longest straight line in an isovist. We propose the following three types of transformation of the open space to extract the dominant lines of visibility.

### 2.1. Network of lines of longest depth

As mentioned earlier, it is important to visualise the structural significance of the different parts of the open space for various reasons e.g. accessibility etc. Existing space syntax experiments rely entirely on the partition of open space based on axial lines which still remain subjective. We demonstrate here a rank and overlap elimination (ROPE) methodology which allows the extraction of a different set of lines, found to be similar to the axial lines in both orientation and properties. The methodology is explained in Figure 3. Figure 4 shows an implementation of the ROPE method for the T-shape. The resultant network of lines of longest depth not only contains information about the dominant axes of the space but it is in fact also a trivial solution to the famous art gallery problem [5][4]. However, as can be seen from the figure, the axes convey little

---

[4] Our solution to the art gallery problem is one of the simplest treatments known to us to the problem.

```
                    ┌─────────┐
                    │  Start  │
                    └────┬────┘
                         ▼
        ┌────────────────────────────────────────┐
        │ Rank all viewpoints based on an isovist │
        │ measure e.g. amount of visible area.    │
        └────────────────┬───────────────────────┘
                         ▼
        ┌────────────────────────────────────────┐
        │ Take the isovist of the highest ranking │
        │ viewpoint and remove all viewpoints     │
        │ from space which overlap this isovist.  │
        └────────────────┬───────────────────────┘
                         ▼
        ┌────────────────────────────────────────┐
        │ Keep the viewpoint and longest line     │
        │ through the viewpoint                   │
        └────────────────┬───────────────────────┘
                         ▼
        ┌─────────────────────────────┐    No    ┌──────┐
        │ Viewpoints remain in space? ├─────────▶│ Stop │
        └────────────┬────────────────┘          └──────┘
                    │ Yes
                    ▼
        ┌────────────────────────────────────────┐
        │ Take the isovist of the next ranking    │
        │ viewpoint remaining in space and remove │
        │ all viewpoints which overlap this isovist.│
        └────────────────────────────────────────┘
```

*Figure 3: Rank and Overlap elimination method for the generation of network of longest depth lines.*

*Figure 4: Network of lines of longest depth in the T-shape space. Note how the ROPE methodology also reveals the minimum number of viewpoints (small dots) required to provide complete visual coverage of the open space.*

information about the structure of the T-shape. The main limitation of the network is that it only guarantees optimal visual coverage but no information about the variation in the shape of the open space.

## 2.2. Ridges of the MDL field

As shown in the last section, at each location/viewpoint the orientation of the MDL indicates the direction with the most *depth* visible from the location. Viewpoints with high MDL will therefore tend to dominate the visibility of surrounding viewpoints. Figure 5a shows a 2.5D plot of the MDL measure for T-shape. It is evident from the figure that the visualisation of the *ridges* of MDL reveals the otherwise unnoticed lines of dominant visibility, which could subconsciously influence our movement in the gallery (e.g. the experience of "hey! What's over there?") as we move around in the space. There are many ways for extracting ridge-like morphological features from surfaces. Figure 5b shows an example of a ridge extraction from the MDL field based on the analysis of the local curvature in the MDL surface at each viewpoint. A detailed treatment on feature extraction is beyond the scope of this article. For more information on curvature-based ridge feature extraction, please refer to Wood and Rana [6]. It must be noted that most of the automated feature extraction methods suffer from various limitations that restrict their use therefore a manual extraction of the extracted ridge lines could be more efficient and accurate in some cases. In fact, a completely manual drawing of the ridge lines is not likely to take too long since the ridge lines are always long and straight.

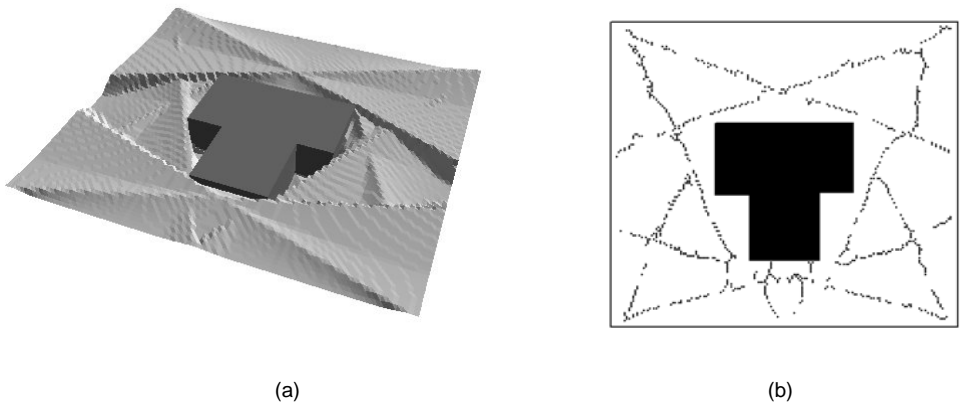

(a)           (b)

*Figure 5: (a) 2.5D plot of the maximum diametric length (MDL) measure. The ridge lines highlight the subtle and explicit clues about the structure of the shape. Also note the merging of ridge lines indicating abrupt changes in vista. (b) Trace of the ridges on the MDL surface after an automated extraction.*

## 2.3. Medial axial lines

Medial axis transform (MAT) or also called skeletonization is a popular approach to abstract the structure of shapes and used in a variety of fields such as optical character recognition and computer-aided design. A medial axis (skeleton) is the loci of centers of bi-tangent circles that fit entirely within a shape being considered [7]. MAT is an image where each point on the skeleton has an intensity which represents its distance to a boundary in the original shape. It is easy to imagine that such a description is very useful for architectural open space however this is relatively unknown in the study of architectural open spaces. A simple method of computing medial axes of a shape is to calculate the distance transform of the shape i.e. calculate the distance to the shape boundaries all over the shape. The medial axis skeleton lies along the *singularities* (*i.e.* creases or curvature discontinuities) in the distance transform [7]. Van Tonder et al. [8] have proposed that medial axis lines pose a strong influence on our appreciation of a scene and hence some of the ancient designers could have taken this into account.

In isovist analysis, the calculation of the MRL is in fact same as the distance transform. Figure 6a shows a 2.5D plot of the MRL measure of the T-shape. Figure 6b shows an extraction of the medial axes (ridges of the MRL surface) using the method given in the last section.

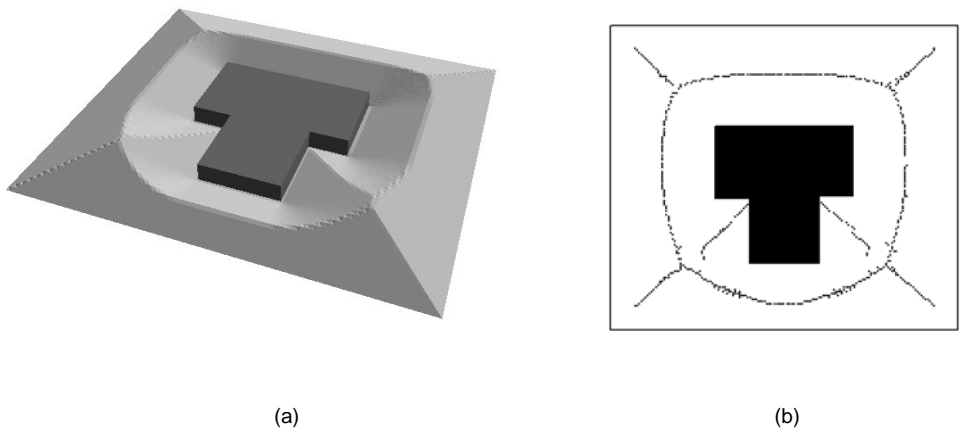

(a)          (b)

*Figure 6: (a) 2.5D plot of the minimum radial length (MRL) measure. The ridge lines demarcate the medial axes and connect together to describe the topological structure. (b) Trace of the medial axes after an automated extraction.*

## 4. CASE STUDY

With these descriptions, it is now possible to describe the structure of real open spaces in cities. In this work, an example of the open space in the central Manchester, UK is presented. Figure 7 shows the footprints of few non-residential buildings in central Manchester and the amount of the open space between them. The layout of the building footprints has a clearly distinguishable structure and the use of our measures will highlight this aspect of the given open space. Figure 8a, 8b, and 8C show the MDL field, the ridges and network of the longest lines of depth derived from the MDL surface respectively, essentially highlighting the long street corridors. Visualisation was obtained almost effortlessly unlike the rigorous analysis of convex subspaces required in the conventional space syntax methods. However, it must be noted that the network of lines of longest depth is not at all same as axial lines but the network provides equivalent information about the structure of open space. One of the most obvious differences between the network of lines of depth and axial lines is the strong influence of minor irregularities of object boundaries on the result. In conventional axial line analysis, these minor irregularities will have to be discarded/aggregated into the bigger subspace. However, the network of lines of longest depth will preserve these features and as a result often multiple lines of longest depth may occupy a single sub-space, as also seen in figure 8c. The benefit of this sensitivity is that apart from visualisation, the network of lines of longest depth can now also be used by police for surveillance e.g. placement of CCTV camera in the town centre, by city council for traffic analysis, accessibility and so on. Figure 9a and 9b show the plot of the MRL field and the corresponding ridges in the surface. Again, the ridges highlight the convex and concave edges in the open space which reveal information about the symmetry of the space for aesthetic purposes and the street layout for more practical applications.

## 5. CONCLUSION

In this paper, we have demonstrated three ways of visualising and abstracting the structure of open spaces based on shape descriptors. However, the descriptors are universal to any type of shapes and hence are equally applicable to most kinds of shape analysis.

Although we have formalised an approach to detect the dominant lines of visibility, it remains to be tested whether the human navigation can be modelled under these constraints. One of the obvious limitations of our methods is that they are based on a 2D open space. However, our appreciation of space is clearly based on three-dimensional cognition and involves other influences such as sound. We intend to follow this in our future work. The nature and implications of this analysis requires a broad platform to evaluate these methodologies. Two real

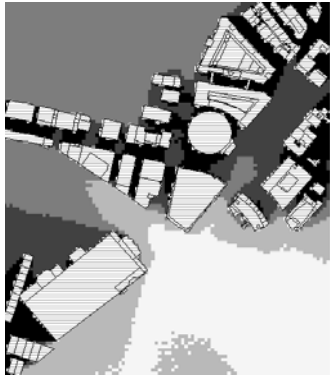

*Figure 7: Non-residential buildings in central Manchester, UK and the distribution of the open space. Lighter colours indicate more visibility and thus more open space.*

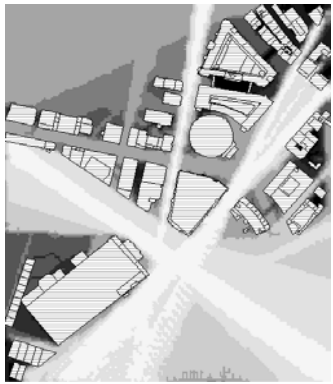
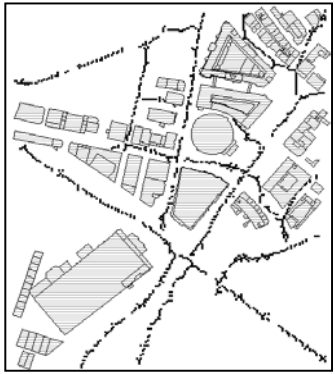
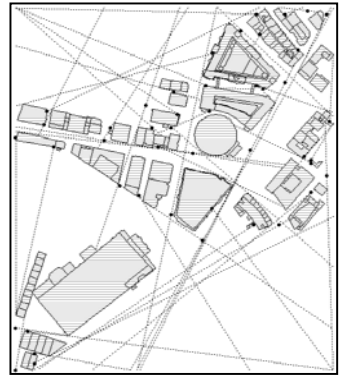

(a)                  (b)                  (c)

*Figure 8: (a) Long corridors between the central Manchester built structure highlighted by the pattern of MDL measure at different locations in the open space. (b) Trace of the ridges on the MDL surface approximately following the long visual corridors. (c) Network of lines of longest depth indicating the key visibly important locations (black dots) and their respective dominant viewing directions shown by orientation of lines of longest depth.*

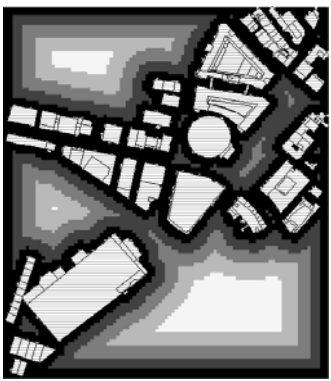
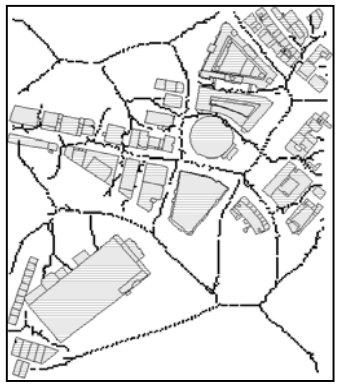

*Figure 9: Skeleton of the open space between the central Manchester shown by MRL pattern. (a) MRL pattern showing the shape of open space. (b) Trace of the ridges of the MRL surface.*

(a)                  (b)

examples of this approach were presented in an earlier paper by us [3]. The choice of the example of central Manchester was deliberate because Manchester is one of the most vibrant cities in England and hence by taking this as an example, we hope that the researchers will be encouraged to experiment with these ideas. We believe that we have mentioned only a few of the many potential applications of our approach.

Finally, it must be stressed that visualisation of an open space in quantifiable terms has to involve some notion of qualitative aspects of human behaviour such as our different appreciation of colour, sound, organisation of objects in space and so on. This remains the topic of our future work especially to derive analogies between the methods and measures of these two forms of visualisation exercise.

**Acknowledgements**

Manchester building footprints were obtained from the Non-Domestic Building Project at University College London. Lakshman Prasad at Las Alamos National Laboratory kindly provided intellectual and materials help during the research of this paper.